# Combining Spatial and Telemetric Features for Learning Animal Movement Models

**Berk Kapicioglu**
Princeton University
Dept. of Computer Science
bkapicio@cs.princeton.edu

**Robert E. Schapire**
Princeton University
Dept. of Computer Science
schapire@cs.princeton.edu

**Martin Wikelski**
Max Planck Institute
for Ornithology
martin@orn.mpg.de

**Tamara Broderick**
University of California, Berkeley
Department of Statistics
tab@stat.berkeley.edu

## Abstract

We introduce a new graphical model for tracking radio-tagged animals and learning their movement patterns. The model provides a principled way to combine radio telemetry data with an arbitrary set of user-defined, spatial features. We describe an efficient stochastic gradient algorithm for fitting model parameters to data and demonstrate its effectiveness via asymptotic analysis and synthetic experiments. We also apply our model to real datasets, and show that it outperforms the most popular radio telemetry software package used in ecology. We conclude that integration of different data sources under a single statistical framework, coupled with appropriate parameter and state estimation procedures, produces both accurate location estimates and an interpretable statistical model of animal movement.

## 1 INTRODUCTION

Animals move through their environments in complex ways. Understanding the processes that govern animal movement is a fundamental problem in ecology and has important ramifications in areas such as home-range and territorial dynamics, habitat use and conservation, biological invasions and biological control [7]. Ecologists rely heavily on collecting and analyzing animal movement data to deepen their understanding of these processes.

In recent years, various technological advances, such as radio telemetry systems and the Global Positioning System (GPS), have created new avenues for collecting data from animals [17]. In radio telemetry, animals are tagged with a tiny radio transmitter. At regular time intervals, fixed-location towers in the environment record the signal from the transmitter and use this information to infer the direction of the animal with respect to the tower. In contrast to GPS, radio telemetry systems can use smaller transmitters, can collect data more frequently, are simpler to implement, and can be used under rainforest canopies. An implementation of such a system in Barro Colorado Island, called the Automated Radio Telemetry System Initiative (ARTS), provides researchers access to hundreds of thousands of directional data measurements collected from a variety of animals [4]. However, even though this method has led to a proliferation of directional data measurements, it has been difficult to harness the full potential of these data sets because: 1) directional measurements obtained through radio telemetry are notoriously noisy; 2) telemetry databases may contain very large amounts of data; and 3) directional measurements are not in a form that is easily interpretable. What is needed are computational tools that would efficiently and accurately convert these large and noisy data sets into a form that would enhance ecological research.

Previously, various sequential probabilistic graphical models have been proposed to solve this problem. Anderson-Sprecher and Lodelter [1, 2] used an iterated extended Kalman filter-smoother to estimate animal locations. Jonsen et al. [6, 7] represented animal movements as correlated random walks and used Bayesian techniques to infer posterior model parameters and animal locations. Ovaskainen [13] used a diffusion approach to model movement in heterogeneous landscapes. Morales et al. [11] fitted multiple random walks to animal movement paths and modeled switching probabilities between them as a function of landscape variates. Patterson et al. [14] and Schick et al. [15] further review past probabilistic graphical models and their applications to ecology and animal movement.

In this work, we propose a new state space model

(SSM) approach. In contrast to previous work, our SSM includes a richer and more general animal movement model. It can utilize arbitrary geographical features, such as the population densities of various tree species or the location of water sources, to represent geographically-dependent animal movement models that cannot be represented by previous approaches at this generality. While our animal movement model also generalizes Gaussian random walks, which are models associated with Kalman filters, combining and incorporating spatial features, especially geographical ones, yields an interpretable model that enunciates the relationship between the animal and its environment. Furthermore, a richer model leads to increased accuracy in estimating animal locations from telemetry data.

In order to handle the challenges of incorporating a very large and feature-rich latent state space, we present an efficient stochastic gradient algorithm for learning the parameters of the model. We demonstrate the algorithm's effectiveness in training our model by comparing it to the expectation-maximization (EM) algorithm both via asymptotic analysis and synthetic experiments. We note that, in contrast to sequential Monte Carlo methods such as particle filters, our inference algorithms allow us to train parameters and infer past estimates based on *all* observations.

Finally, we apply our model to real datasets and demonstrate that it outperforms the most popular radio telemetry software package used in ecology. In conclusion, we show that our methods aid us in producing interpretable results and accurate animal location estimates from large and noisy telemetry datasets.

This paper is organized as follows. In Section 2, we specify the SSM and formalize its parameter and state estimation problems. In Section 3, we detail the EM and Viterbi algorithms for solving the estimation problems and analyze their time-complexity. In Section 4, we present and analyze the stochastic gradient algorithm. We report the results of our experiments on synthetic and real datasets in Section 5 and conclude in Section 6.

## 2 A STATE SPACE MODEL

In this work, we define a telemetry data set as a long sequence of directional measurements that are observed by a small number of fixed-location towers at regular time intervals. Note that all directional measurements in a given telemetry data set are received from the same animal. We model the data as if it were generated by an SSM. In this section, we describe the SSM by detailing its latent state space as well as its start, transition and observation models. Then, we formalize the parameter and state estimation problems, and argue how solving the estimation problems yields an interpretable model of animal movement and accurate animal location estimates.

Intuitively, the latent state space is the space of possible animal locations. In order to use the SSM machinery, we discretize a continuous latent state space of animal locations into a finite and discrete space of coordinates $Q \subset \mathbb{R}^2$. In practice, $Q$ is constructed by partitioning a larger state space into finitely many equally sized grid cells and assigning each midpoint of a grid cell as a coordinate in $Q$.

The start model generates the first element of the latent animal location data. For simplicity's sake, we use the uniform distribution as the start model. Let the latent random state at time step $t \in \{1, \ldots, T\}$ be denoted by $x_t \in Q$. Then, the start model is:

$$p(x_1) = \frac{1}{|Q|}. \quad (1)$$

The transition model, a Gibbs distribution, generates the rest of the latent animal location data. Gibbs distributions, also known as the conditional exponential model, have emerged as popular models in machine learning due to their practical success and their theoretical elegance. These distributions are defined using features, which in our case are functions that encode information about the spatial properties of the environment. For example, a feature may encode the distance between two coordinates in the latent state space, or it may encode the minimum distance to a certain tree species. In a slightly different but related setting, it has been shown that the maximum likelihood estimate of a Gibbs distribution is equivalent to the maximum entropy distribution with various constraints imposed by the features [3]. In our case, we adopt the Gibbs distribution as the transition model because it provides a means to incorporate spatial features without making any independence assumptions about them, remains resilient to the extension of feature space by irrelevant features, and generalizes discrete versions of simpler random walk models which have previously been used to model animal movement.

More formally, let $f_k : Q \times Q \to \mathbb{R}$ denote the $k$th feature and $\lambda_k \in \mathbb{R}$ denote the corresponding weight parameter, where $k \in \{1, \ldots, K\}$. We use $\boldsymbol{\lambda}$ as a vector representation of feature weights. Then the transition model is:

$$p(x_{t+1}|x_t; \boldsymbol{\lambda}) = \frac{\exp\left(\sum_{k=1}^{K} \lambda_k f_k(x_t, x_{t+1})\right)}{\sum_{x_q \in Q} \exp\left(\sum_{k=1}^{K} \lambda_k f_k(x_t, x_q)\right)}. \quad (2)$$

The observation model generates the directional measurements observed by the towers. We model the behavior of towers as a von Mises distribution, a circular analogue of the normal distribution [10]. A von Mises distribution is parameterized by $\mu$ and $\kappa$, roughly the analogs of the mean and the variance of a normal distribution, respectively. Intuitively, a radial bearing is sampled from a von Mises distribution, and it is added as noise to the true bearing that points in the direction of the animal.

More formally, fix a radial bearing system shared by all towers (i.e., bearing $\frac{\pi}{2}$ points north and bearing 0 points west). Let $y_{t,n} \in [-\pi, \pi)$ denote the random variable, a radial bearing, observed by tower $n \in \{1, \ldots, N\}$ at time step $t$. This is a noisy observation that is supposed to point in the direction of the animal. Let $z_n \in \mathbb{R}^2$ denote the coordinate of tower $n$ and define $h(x, z_n) \in [-\pi, \pi)$ to be the true radial bearing of the vector that points from tower $n$ towards location $x$. Let $\mu_n \in \mathbb{R}$ and $\kappa_n \geq 0$ denote the parameters of the von Mises distribution for tower $n$. Let $I_0$ denote the modified Bessel function of the first kind with order 0, which simply acts as the normalization constant for the von Mises distribution. Finally, let $\boldsymbol{\mu}$, $\boldsymbol{\kappa}$, and $\mathbf{y}_t$ be the vector representations of the corresponding parameters and random variables. Then, the observation model is:

$$p(\boldsymbol{y}_t | x_t; \boldsymbol{\mu}, \boldsymbol{\kappa}) = \prod_{n=1}^{N} p(y_{t,n} | x_t; \mu_n, \kappa_n)$$
$$= \prod_{n=1}^{N} \frac{\exp(\kappa \cos(y_{t,n} - h(x_t, z_n) - \mu_n))}{2\pi I_0(\kappa)}. \quad (3)$$

The factorization occurs because of the conditional independence assumptions that hold between the observations.

The start, transition, and observation models can be used to compute all the marginal and conditional probability distributions of the SSM. Let $\mathbf{y}$ denote the vector of all observations and let $\boldsymbol{\theta} = (\boldsymbol{\lambda}, \boldsymbol{\mu}, \boldsymbol{\kappa})$ denote the vector of model parameters. As usual, the joint probability distribution of the SSM is:

$$p(\boldsymbol{x}, \boldsymbol{y}; \boldsymbol{\theta}) = p(\boldsymbol{x}, \boldsymbol{y}; \boldsymbol{\lambda}, \boldsymbol{\mu}, \boldsymbol{\kappa})$$
$$= p(x_1) \prod_{t=1}^{T-1} p(x_{t+1} | x_t; \boldsymbol{\lambda}) \prod_{t=1}^{T} p(\boldsymbol{y}_t | x_t; \boldsymbol{\mu}, \boldsymbol{\kappa}). \quad (4)$$

Parameter estimation is the problem of estimating the model parameters (i.e. $\boldsymbol{\theta}$). We choose maximum likelihood estimation (MLE) as the method of fitting model parameters to data. We define the problem of parameter estimation as:

$$\hat{\boldsymbol{\theta}} = \left(\hat{\boldsymbol{\lambda}}, \hat{\boldsymbol{\mu}}, \hat{\boldsymbol{\kappa}}\right) = \underset{\boldsymbol{\lambda} \in \mathbb{R}^K, \boldsymbol{\mu} \in \mathbb{R}^N, \boldsymbol{\kappa} \geq \mathbf{0}}{\arg\max} \log p(\mathbf{y}; \boldsymbol{\lambda}, \boldsymbol{\mu}, \boldsymbol{\kappa}). \quad (5)$$

State estimation, on the other hand, is the problem of estimating values of unobserved random variables. We choose to seek the hidden state sequence that maximizes the probability conditioned on the observations and the MLE parameter estimates. More formally, we define the problem of state estimation as:

$$\hat{\mathbf{x}} = \underset{\mathbf{x} \in Q^T}{\arg\max} \, p\left(\mathbf{x} | \mathbf{y}; \hat{\boldsymbol{\theta}}\right). \quad (6)$$

Our intention is to use the parameter estimates of the Gibbs distribution as an interpretable model of animal movement, and the state estimates of the latent random variables as animal location estimates. We note that an analysis of Gibbs weights might not conclusively explain the processes governing animal movement, but we hope that it will be a first step in further exploration of these processes.

## 3 EM

In this section, we detail the EM and Viterbi algorithms for solving the estimation problems, and analyze their time-complexity. Later in the paper, we will see that for the type of datasets we have, where both the cardinality of the latent state space and the number of time steps is large, but the cardinality of the feature space is small, the stochastic gradient algorithm is asymptotically superior to EM. The analysis in this section will allow us to compare EM and the stochastic gradient algorithms.

---
**Algorithm 1** EM( $\boldsymbol{\theta}^0$, $maxTime$)
---
1: $i \leftarrow 0$.
2: **repeat**
3:     $\forall t \in \{1, \ldots, T\}$ and $\forall x_t \in Q$, compute $\log p(x_t | \mathbf{y}; \boldsymbol{\theta}^i)$.
4:     $\forall t \in \{1, \ldots, T-1\}$ and $\forall (x_t, x_{t+1}) \in Q^2$, compute $\log p(x_t, x_{t+1} | \mathbf{y}; \boldsymbol{\theta}^i)$.
5:     $\forall n \in \{1, \ldots, N\}$, $(\mu_n^{i+1}, \kappa_n^{i+1}) \leftarrow$
$$\underset{\mu_n, \kappa_n}{\arg\max} \sum_{t=1}^{T} \underset{x_t | \mathbf{y}; \boldsymbol{\theta}^i}{\mathbb{E}} \left[\log p(y_{t,n} | x_t; \mu_n, \kappa_n)\right].$$
6:     $\boldsymbol{\lambda}^{i+1} \leftarrow \underset{\boldsymbol{\lambda}}{\arg\max} \sum_{t=1}^{T-1} \underset{x_t | \mathbf{y}; \boldsymbol{\theta}^i}{\mathbb{E}} \left[\log p(x_{t+1} | x_t; \boldsymbol{\lambda})\right].$
7:     $i \leftarrow i + 1$.
8: **until** elapsed time exceeds $maxTime$.
9: **return** $\boldsymbol{\theta}^i$.

---

In Algorithm 1, we present the adaptation of EM to our setting. Here, $maxTime$ denotes the maximum allowed running time and $\boldsymbol{\theta}^0 = (\boldsymbol{\lambda}^0, \boldsymbol{\mu}^0, \boldsymbol{\kappa}^0)$ denotes the initial parameter settings. Lines 3 and 4 describe what is traditionally called the E-step and involve computations of various conditional log-probabilities. These log-probabilities can be computed using algorithms such as the forward-backward algorithm. Similarly, Lines 5 and 6 describe what is traditionally called the M-step and involve maximization of expected complete log-likelihoods. The maximization problem in Line 5, after substitution of Equation (3) and some trigonometric manipulations, can be rewritten as:

$$\arg\max_{\mu_n,\kappa_n} \left( \sum_{t=1}^{T} \kappa_n \cos \mu_n \mathop{\mathbb{E}}_{x_t|\boldsymbol{y};\boldsymbol{\theta}^i} [\cos(y_{t,n} - h(x_t, z_n))] \right)$$
$$+ \left( \sum_{t=1}^{T} \kappa_n \sin \mu_n \mathop{\mathbb{E}}_{x_t|\boldsymbol{y};\boldsymbol{\theta}^i} [\sin(y_{t,n} - h(x_t, z_n))] \right)$$
$$- T \log 2\pi I_0(\kappa_n). \quad (7)$$

In this form, it is equivalent to the problem of finding MLE estimates of von Mises parameters from i.i.d samples, and it can be solved in closed form using techniques from directional statistics [10]. The maximization problem in Line 6 is convex and unconstrained, and it can be solved using numerical optimization algorithms such as BFGS [9, 12].

EM is an iterative algorithm and it is hard to predict beforehand how many iterations are necessary until convergence. Here, we analyze the time-complexity of each iteration for a *direct* implementation of Algorithm 1. When it is clear from context, we write $Q$ for $|Q|$. We also remind the reader that $Q$ denotes the latent state space, $K$ denotes the number of features, and $T$ denotes the number of time steps. Almost all EM computations rely on the transition matrix, which can be computed in $\Theta(Q^2 K)$. The E-step, Lines 3 and 4, consists of computation of conditional log probabilities, which takes $\Theta(Q^2 T)$. The M-step, Lines 5 and 6, consists of both iterative and non-iterative optimization algorithms for estimating Gibbs and von Mises parameters, respectively. Optimization of von Mises parameters can be computed in $\Theta(QT)$, whereas optimization of Gibbs parameters is itself conducted iteratively using BFGS. However, in practice, the number of BFGS iterations is small. For each BFGS iteration, it takes $\Theta(Q^2 T)$ to evaluate the objective function and $\Theta(Q^2 K)$ to evaluate its gradient. Finally, we note that we suppressed the dependence on $N$, the number of towers, since it is small. Overall, each EM iteration takes $\Theta(Q^2 (K + T))$.

For solving the problem of state estimation, Equation (6), the Viterbi algorithm may be used. Viterbi is not iterative and it only needs to be executed once. Once the transition matrix is computed in $\Theta(Q^2 K)$, the most likely path is computed in $\Theta(Q^2 T)$, yielding, like EM, a total running time of $\Theta(Q^2 (K + T))$.

An important factor in determining both computational efficiency and statistical accuracy is the resolution of the grid used to construct $Q$, the latent state space. As the diagonal distance between corners of each grid cell in $Q$ gets shorter, the latent state space representation becomes more accurate[1] linearly, whereas the cardinality of the latent state space increases quadratically. Thus, an increase in statistical accuracy at a linear rate is offset by an increase in computational complexity at a *quartic* rate, since EM and Viterbi themselves have time-complexity quadratic in the cardinality of the latent state space. This trade-off leads to a rapid increase in the running time of these algorithms in exchange for small gains in statistical accuracy.

## 4 STOCHASTIC GRADIENT

In this section, we propose a stochastic gradient alternative to EM, which under certain conditions, is asymptotically superior to EM. The algorithm is an optimization method that can be used when one has access to a noisy approximation of the gradient of the objective function [16]. Younes [18] used it to estimate parameters of partially observed Markov Random Fields. Delyon et al. [5] proved convergence results for a family of stochastic approximations to EM, including the stochastic gradient algorithm of the form depicted in Algorithm 2. In the rest of the section, we present our implementation of the stochastic gradient algorithm, analyze its time-complexity and compare it to that of EM, and detail how our implementation differs from other implementations.

---

**Algorithm 2** SG( $\boldsymbol{\theta}^0$, $numBurn$, $maxTime$)

1: $i \leftarrow 0$.
2: **repeat**
3:   Sample $\boldsymbol{x}^i \sim p(.|\mathbf{y};\boldsymbol{\theta}^i)$ using randomized Gibbs sampling with a burn-in period of $numBurn$.
4:   $\boldsymbol{\theta}^{i+1} \leftarrow \boldsymbol{\theta}^i + \gamma_i \nabla L_{\boldsymbol{x}^i}(\boldsymbol{\theta}^i)$, where $\gamma_i$ is chosen to satisfy Wolfe conditions with respect to $L_{\boldsymbol{x}^i}(\boldsymbol{\theta}^i)$.
5:   $i \leftarrow i + 1$.
6: **until** elapsed time exceeds $maxTime$.
7: **return** $\boldsymbol{\theta}^i$.

---

The algorithm is presented formally as Algorithm 2.

---
[1] One measure of accuracy is the diagonal distance between the corners of a grid cell. In the worst-case scenario, a larger grid cell would yield a higher absolute error between the location estimate and the true location even when the true location is estimated by the closest cell midpoint.

Here, $\boldsymbol{\theta}^0$ denotes the initial parameter settings, $maxTime$ denotes the maximum allowed running time, and $numBurn$ denotes the number of Markov Chain Monte Carlo (MCMC) iterations executed before a sample obtained through Gibbs sampling is accepted. The log-likelihood of the complete data, denoted $L$, is defined as:

$$L_{\mathbf{x}}(\boldsymbol{\theta}) = L_{\mathbf{x}}(\boldsymbol{\lambda}, \boldsymbol{\mu}, \boldsymbol{\kappa}) = \log p(\mathbf{x}, \mathbf{y}; \boldsymbol{\lambda}, \boldsymbol{\mu}, \boldsymbol{\kappa}). \quad (8)$$

In Line 3, latent random variables are sampled from the conditional distribution using Gibbs sampling, and in line 4, model parameters are updated in the direction of the gradient. We note that the algorithm is a stochastic gradient algorithm because the gradient of the log-likelihood of the complete data, Equation (8), is a noisy approximation of the gradient of the objective function, Equation (5).

The main difference between our implementation of the stochastic gradient algorithm and previous implementations is the choice of the learning rate $\gamma$. Common implementations of the stochastic gradient algorithm use a decreasing learning rate sequence that guarantees the convergence of the algorithm to a stationary point. [5]. However, in practice, the choice of the constant factor associated with such sequences significantly influences the speed of convergence and it is a difficult task to identify the optimal factor. We tune the learning rate automatically by choosing one that satisfies the Wolfe conditions. These are line search conditions that guarantee the convergence of gradient updates to an optimum in *deterministic* convex optimization settings [12], but in our case, we use them in a *stochastic* optimization setting. In practice, parameters that satisfy these conditions can be found by providing the objective function and its gradient to line search methods. Even though we lack a proof of convergence, we have observed informally that our criterion for selecting the learning rates ensures better performance than manually choosing learning rates that satisfy the customary assumptions. Furthermore, our approach provides an automated way to set the learning rate based on the dataset.

We conclude the section with an analysis of the time complexity of our implementation of Algorithm 2. Line 3 takes $\Theta(Q(B + VK))$, where $B$ is the burn-in period and $V$ is the number of unique states visited during sampling. In order to achieve this complexity, we compute the transition probabilities only for the sampled states, and use memoization to store and recall them. Line 4 is conducted iteratively, due to the line search that involves finding the right learning rate, but number of iterations is very small in practice. Computing the objective function takes $\Theta(QVK + T)$, and computing the gradient takes $\Theta(K(QV + T))$.

In conclusion, the choice between EM and the stochastic gradient algorithm depends on the relative size of the latent state space, the feature space, and the number of time steps; however, in our problem, where both the cardinality of the latent state space and the number of time steps is large, but the cardinality of the feature space is small, the stochastic gradient algorithm is asymptotically superior.

An asymptotic comparison of each iteration does not suffice to determine which algorithm to use; thus in the next section, we perform empirical tests that compare these algorithms with respect to a variety of performance measures. Also, we note that *both* algorithms may be optimized such that only state transitions to a small neighborhood are taken into account in the computations. This would reduce time-complexity from $\Theta(Q^2)$ to $\Theta(QR)$, where $R$ is the cardinality of the largest neighborhood. Our implementation of all discussed algorithms take advantage of this optimization.

## 5 EXPERIMENTS

In this section, we demonstrate our model on both synthetic and real datasets. First, we generate synthetic datasets to evaluate our model in a setting where true feature weights are known. We also use the synthetic datasets to compare the performance of the EM and the stochastic gradient algorithms. Then we use the real datasets to compare our model with LOAS, the most ubiquitous radio telemetry software package used in ecology.

### 5.1 SYNTHETIC DATASETS

In order to simulate a real-world application as closely as possible, we created a virtual island that has approximately the same dimensions and the same tower locations as the Barro Colorado Island. We partitioned the virtual island into 3654 grid cells (latent states), where each grid cell is $100 \times 100$ meters square. Then we created 10 different animal movement models (transition models). Each animal movement model consisted of 5 features whose function values were generated uniformly at random from $[0, 1)$, and each feature's weight was generated uniformly at random from $[-10, 10]$. For each animal movement model, we generated an animal path of length 1000. In order to be as realistic as possible, we constrained the animal to move at most 500 meters at each time step and normalized the probabilities of the animal movement model accordingly. We also used the same constraint during the parameter estimation. As for the von Mises parameters of the towers, we set each $\mu$ to 0 and $\kappa$ to 15, which approximates a normal distribution with a standard deviation of 15 degrees. For each of the 10 animal paths, we generated the corresponding noisy bearings, and together with the corresponding features (but without the feature weights), provided them as input to the parameter estimation algorithms.

In the first batch of experiments, we compared the performance of the stochastic gradient algorithm and the EM algorithm. For both algorithms, we set their initial Gibbs weights to 0, which corresponds to a uniform transition model, and their initial $\kappa$ parameters to 50. For the stochastic gradient algorithm, we used a burn-in period of $100,000$. We executed the algorithms on each dataset for 10 hours. To evaluate their performance, we measured the arithmetic mean error of the location estimates, the Euclidean distance between the learned weights and the true weights, and the log-likelihood of the observed bearings. We used the Euclidean distance between the weights as a measure of the interpretability of the model, where the closer weights were considered more interpretable.

As suggested by the asymptotic analysis, the stochastic gradient algorithm executed many more iterations than EM, and outperformed EM with respect to both the arithmetic mean error of the location estimates and the Euclidean distance between the learned weights and the true weights. The stochastic gradient algorithm also attained a higher log-likelihood than EM. We display the results, averaged over the 10 datasets, in Figure 1.

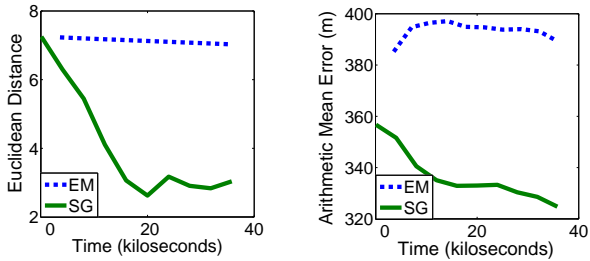

Figure 1: In these plots, we compare the performance of the EM and the stochastic gradient algorithms. During 10 hours, EM had an average of 10 iterations, whereas stochastic gradient had an average of 500 iterations. The left plot reports the Euclidean distance between the learned weights and the true weights, and the right plot reports the average mean error of the location estimates. Stochastic gradient outperforms EM in both cases.

In the second batch of experiments, we compared the performance of our animal movement model with a discrete version of the animal movement model used by Anderson-Sprecher [1]. By doing so, we both wanted to demonstrate how our model can generalize previous animal movement models and we wanted to observe whether having a richer model leads to an improvement in the accuracy of location estimates. The animal movement model used by Anderson-Sprecher is an isotropic bivariate Gaussian random walk which was trained using an extended Kalman filter-smoother. His model is defined as:

$$p\left(x_{t+1}|x_t;\sigma^2\right) = \frac{1}{2\pi\sigma^2} \exp\left(-\frac{\|x_{t+1}-x_t\|^2}{2\sigma^2}\right). \quad (9)$$

We can represent the same model approximately as a Gibbs distribution using the single distance-based feature $f_{dist} = -\frac{\|x_{t+1}-x_t\|^2}{2}$. Similar to the first batch of experiments, we executed two copies of the stochastic gradient algorithm on the same datasets; one copy used the features that generated the datasets and the other copy used the single feature that represents the bivariate Gaussian random walk. The richer model outperformed the Gaussian random walk model with respect to the accuracy of the location estimates. We display the results of these experiments, averaged over 10 datasets, in Figure 2.

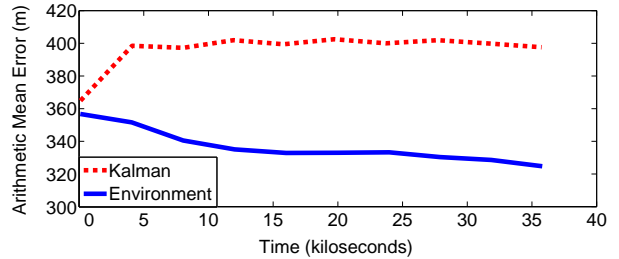

Figure 2: In this plot, we compare the performance between using a richer feature-based (Environment) model versus using a simpler random walk (Kalman) model. The results imply that, if the animal indeed moves around based on environmental features, a model that incorporates such features does yield better location estimates than a simpler random walk model.

## 5.2 REAL DATASETS

We applied our model to the radio telemetry data collected from two sloths, named Chispa and Wendi, that live on the Barro Colorado Island. These sloths were one of the few animals on the island whose true locations were labeled every few days by human researchers via GPS devices. Thus, we were able to use the radio telemetry data to train our algorithms and the GPS data to test them.

The radio telemetry data was collected every 4 minutes for 10 days, yielding approximately 3600 time points. Both datasets had considerable noise in the bearing measurements. In order to apply our model, we discretized the island into 1200 grid cells, each of size $50 \times 50$ meters square. As features, we used both the distance-based feature that encodes a bivariate random walk, Equation (9), and tree-based features that encode the change in the population density of various tree species across grid cells. In particular, the model included 18 features: 1 distance-based feature and 17 tree-based features. Each of the 17 tree-based features corresponded to a different tree species and they were normalized to have values in $[-1, 1]$. We initialized the Gibbs weights to 0, the $\mu$ parameters to 0, and the $\kappa$ parameters to 10. We executed 10 different instances of the stochastic gradient algorithm, each using a different random seed, and averaged the results. As a baseline comparison, we used LOAS, which is the most popular radio telemetry software package used in ecology [8].

We display the results in Figure 3 and Figure 4. For both datasets, SSM outperformed LOAS with respect to the accuracy of the location estimates. As demonstrated for the Wendi dataset, our model successfully learned a diminishing movement variance for the sloth, represented in the top-right figure as the growing weight associated with the distance-based feature. It also identified a small portion of the tree species that the sloth seems to have a preference for.

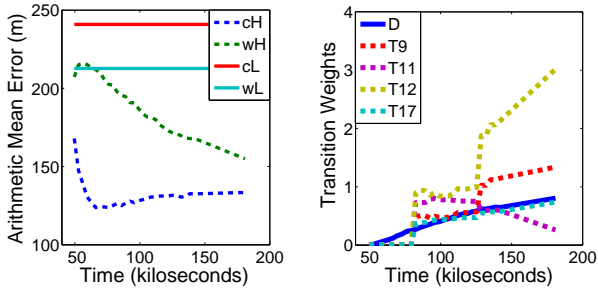

Figure 3: **Left**: This plot compares the location estimation performance of the two algorithms. cH and wH represent the performance of the SSM; cL and wL represent the performance of LOAS. The initial letters "c" and "w" refer to the Chispa and Wendi datasets, respectively. **Right**: This plot displays the evolution of the transition weights for the Wendi dataset. We only plotted the weights that exceeded the 0.5 threshold. "D" and "T" denote the distance-based and tree-based features, respectively.

We also generalized our movement model features to be temporally-dependent. In particular, we partitioned each day into different time zones, and for each such zone, we allowed the animals to move according to a different movement model. The changes in the daily activity of the sloths have already been studied by field biologists; so for our experiments, we set the time zones, "day" and "night", based on biologists' feedback.

In our last experiment, we were interested in evaluating whether feature weights converge to meaningful values, whether they are interpretable, and how different types of features interact with each other. For this purpose, we created an *artificial* tree type, where for each of the very few locations GPS data was available for Wendi, we placed a virtual tree. We defined this tree-feature formally as $f_{tree} = -\frac{\|tree\_dist(x_{t+1}) - tree\_dist(x_t)\|^2}{2}$, where $tree\_dist(x)$ is the Euclidean distance between $x$ and the closest tree of that type. By defining the tree-based feature this way, we set it to the "same" unit as the distance-based feature associated with Equation (9), allowing the feature weights to be numerically comparable. After normalizing all feature values to be within $[-1, 1]$, the weights converged to 0.704 for distance-based feature at night and 0.316 during day; $-1.279$ for tree-based feature at night and $-1.312$ during day. As expected, interpreting distance-based weights indicate that Wendi moves more during day than night and that the positive magnitude of these weights indicate that the animal is more likely to stay in place in succeeding time steps. Also, tree-based weights correctly indicate that Wendi has a strong preference for certain neighborhoods on the map (i.e. the true locations), and that the strength of these preferences are indifferent to the time zone.

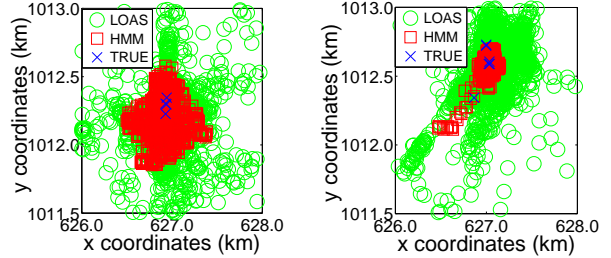

Figure 4: These plots display the true locations, SSM estimates, and LOAS estimates. SSM estimates are based on the last stochastic gradient iteration. Left plot displays the results for the Wendi dataset and the right plot displays the results for the Chispa dataset.

## 6 CONCLUSION

In this paper, we presented an SSM approach to locate radio-tagged animals and learn their movement patterns. We presented a model that incorporates both geographical and non-geographical spatial features to improve animal location estimates and provide researchers an interpretable model that enunciates the relationship between the animal and its environment. We showed that the model generalizes discrete versions of random walk models and demonstrated empirically that a richer model improves animal location estimates. We also provided a fast parameter estimation algorithm, the stochastic gradient algorithm, and demonstrated its effectiveness against

EM both asymptotically and empirically. Finally, we applied our model to real datasets. Our model outperformed LOAS, the most popular radio telemetry software package in ecology, with respect to the accuracy of the location estimates. Our model also hypothesized that the sloth has a relatively strong preference for certain types of trees. We are currently working on applying our model to other telemetry datasets collected at the Barro Colorado Island.

**Acknowledgements**

This research was supported by the National Science Foundation under Grant #0325463, Smithsonian Tropical Research Institute, Peninsula Foundation, and the National Geographic Society. We thank Roland Kays for his invaluable help in providing us with BCI datasets and answering all our biology-related inquiries. Thanks also to the anonymous reviewers for very constructive suggestions.